\documentclass{svproc}
\usepackage{graphicx}
\usepackage{amsmath}
\usepackage{amssymb}
\usepackage{mathtools}
\usepackage{multirow}
\usepackage{stmaryrd}
\usepackage{multirow}
\usepackage{etoolbox}

\usepackage[caption=false]{subfig}
\usepackage[hyphens]{url}
\usepackage[breaklinks]{hyperref}
\usepackage{booktabs} 
\usepackage[bottom]{footmisc}

\def\ptFiguresDirectory#1{./#1}
\def\FWER#1{FWER}
\def\fdr{\mathrm{FDR}}
\def\fnr{\mathrm{FNR}}
\def\fOne{F_{1}}

\def\normSDParam{\gamma}

\newcommand{\argmax}{\arg\!\max}

\begin{document}
\frontmatter          
\pagestyle{headings}  
\addtocmark{} 
\mainmatter              
\title{Randomized Reference Classifier with Gaussian Distribution and Soft Confusion Matrix Applied to the Improving Weak Classifiers}
\titlerunning{RRC with Gaussian Distribution}  
%
\author{Pawel Trajdos\inst{1} \and Marek Kurzynski\inst{1}}
\authorrunning{Pawel Trajdos \and Marek Kurzynski} 
%
\tocauthor{Pawel Trajdos, Marek Kurzynski}
\institute{Wroclaw University of Science and Technology, Wroclaw, Poland,\\
\email{pawel.trajdos@pwr.edu.pl}, \email{marek.kurzynski@pwr.edu.pl}}

\maketitle              

\begin{abstract}
In this paper, an issue of building the RRC model using probability distributions other than beta distribution is addressed. More precisely, in this paper, we propose to build the RRR model using the truncated normal distribution. Heuristic procedures for expected value and the variance of the truncated-normal distribution are also proposed.  The proposed approach is tested using SCM-based model for testing the consequences of applying the truncated normal distribution in the RRC model. The experimental evaluation is performed using four different base classifiers and seven quality measures. The results showed that the proposed approach is comparable to the RRC model built using beta distribution. What is more, for some base classifiers, the truncated-normal-based SCM algorithm turned out to be better at discovering objects coming from minority classes. 

\end{abstract}

\section{Introduction}
\label{sec:Introduction}
Classification, in which one tries to assign a class label to an object, is one of the more common and well-known decision-making problems. Classification or pattern recognition tasks have been successfully applied in many areas including medicine, economy, agriculture, astronomy or defence. In the modern world, there is practically no field of human activity, where computer (automatic) classification methods would not be used. The large practical demand for computer-aided classification algorithms has resulted in the active development of object recognition methodologies in the last few decades. 

Unfortunately, this variety of recognition methods does not always mean an acceptable quality of classification because each problem requires its individual approach and there is no easy solution which classifiers or algorithm should be applied. 

If the built-in classifier does not meet the requirements of the quality of classification or is simply a weak classifier (is a bit better than the random guessing) one can use methods to improve the quality of classifiers.  
The most-known methods are the techniques related to the construction of an ensemble of classifiers on the basis of different training sets created from the original training set in the resampling process with a uniform distribution (bagging) \cite{breiman1996} or uneven distribution and additionally adaptively changed (boosting) \cite{freund1997decision}.

In shortly, bagging applies sampling with replacement to obtain independent training datasets for each individual classifier. Boosting modifies the input data distribution processed by each classifier in a sequence from the results of classifiers trained before, paying more attention to difficult samples. Both algorithms -- even if they were applied to a weak classifier -- lead to a powerful classifier in the form of a multiclassifier system.  

In \cite{MM:Majak2017} authors  introduced algorithm called Bayes metaclassifier (BMC) as a method for improving weak classifier in terms of its classification performance. In general, BMC constitutes the probabilistic generalization of any base classifier independent of its design paradigm and has the form of the Bayes scheme. Since BMC provides probabilistic interpretation for base classifier correct classification and misclassification, this method can be used in sequential classification or as a fusing mechanism in MC systems \cite{MM:Majak:2016}, \cite{MM:MBC:2016}.

In \cite{Trajdos2016} the original method of improving weak classifier was proposed which is based on the concept of soft confusion matrix (SCM), built using validation set. Soft confusion matrix gives a  picture of local properties (for a given test object $x$) of base classifier including empirical probabilities of class-dependent correct and incorrect classifications. The high value of class dependent correct (incorrect) classification probability for a given object denotes that classifier is capable of the correct classification of the object $x$ coming - let say - from the $i$th class (it tends to misclassify object $x$ from  $i$th class to - lets say -  $j$th class). This knowledge can be directly used to correct classifying functions of the base classifier and to improve its quality. 

The developed method additionally requires the formal procedure for calculation of the probability of correct and misclassification of base classifier at the point $x$. For this purpose, the concept of randomized reference classifier (RRC) was used, which originally was proposed in \cite{Woloszynski2011} as a method for calculation of competence of base classifier in the combining procedure of multiclassifier systems. This approach assumes, that the most natural measure of the classifier's competence at a given point of feature space is its probability of correct classification at this point.  
Unfortunately, this probability is equal to 1 or 0, unless we adopt a probabilistic model of the recognition task or assume that the classifier works in a random manner. However, both cases are difficult to accept. First, the competence should be neutral to the base classifier models, and many concepts of classifiers use the probabilistic approach. Second, in MC systems, deterministic base classifiers are generally used. For these reasons, the authors developed an indirect method. In the proposed approach the base classifier is modelled by a hypothetical classifier called randomized reference classifier (RRC). The RRC is defined by a set of random variables, which observed values are class support produced for the object to be classified. Since expected values of random variables are equal to the supports produced by the modelled base classifier, the RRC can be considered - on average - as the equivalent of this classifier. Consequently, the probability of correct classification of RRC at any point of feature space can be used as the competence of modelled base classifier at this point. 
The concept of RRC  proved to be very effective, as it enabled the construction of MC systems, which in experimental research outperformed different state-of-the-art methods. It also turned out that with the help of RRC it is possible to determine other properties of modelled classifiers: the class-dependent probability of correct/incorrect classification and diversification of two classifiers \cite{lysiak2014optimal}, \cite{Trajdos2016}, \cite{woloszynski2009new} and these  RRC capabilities were used in the improved of weak classifier via SCM concept. 

The key problem in the construction of RRC is the choice of the probability distribution of random variables. This choice is not unique and the values of probability of correct classification of RRC and consequently the probabilities of correct/incorrect classification of modelled classifier depend on the definition of the distributions. 
In the original proposition of the RRC, beta probability distributions have been used. Such a choice follows from the specific definition of the class supports produced by the RRC as a random division of the unit interval. From the theory of order, statistics results that then the supports must be beta distributed \cite{David2003}. 
As it seems the use of beta distribution in the draw process is the main disadvantage of the RRC concept.  The only justification, related to the geometrical interpretation of the support vector is not a substantive justification.  In particular, the proposed distribution does not strictly associate the RRC with the modelled base classifier in the context of its properties which are observable for the validation objects. 

In this study, the concept of RRC based on Gaussian distribution is developed. 
Because in Gaussian distribution we can tune not only the expected value as in the beta distribution but also the variance, therefore the proposed RRC classifier is more flexible and can better adapt to the properties of the modelled base classifier.

The paper is divided into four sections and organized as follows. Section~\ref{sec:PropMethod} introduces the formal notation used in the paper and provides a description of the proposed approach. The experimental setup is given in section~\ref{sec:ExSetup}.  In section~\ref{sec:ResAndDisc} experimental results are given and discussed. Section~\ref{sec:Conc} concludes the paper. 

\section{Proposed Method}\label{sec:PropMethod}

\subsection{Preliminaries}\label{sec:PropMethod:Prel}

In the single-label classification approach, a~$d-\mathrm{dimensional}$ vector $\vec{x}\in\mathbb{X}=\mathbb{R}^{d}$ is assigned a~class $m \in\mathbb{M}$, where $\mathbb{M}=\left\{0,1,2,\cdots,M\right\}$ is a set of available classes. The classifier $\psi: \mathbb{X}\mapsto \mathbb{M}$ is an approximation of an unknown mapping $f: \mathbb{X}\mapsto\mathbb{M}$ which assigns the classes to the instances.  The classification methods analyzed in this paper follow the statistical classification framework. Hence, a~feature vector $\vec{x}$ and its label $m$ are assumed to be realisations of random variables ${\vec{\textbf{X}}}$ and ${\vec{\textbf{M}}}$, respectively. The random variables follow the joint probability distribution $P(\vec{\textbf{X}},\vec{\textbf{M}})$. Given the zero-one loss, the optimal decision is made using the maximum \textit{a posteriori} rule:
\begin{align}
 \psi^{*}(\vec{x}) &=\argmax_{\vec{k}\in\mathbb{M}}P(\mathbb{M}=k|\mathbb{X}=x),
\end{align}
where $P(\mathbb{M}=k|\mathbb{X}=x)$ is the conditional probability that the object $x$ belongs to class $k$.

In this paper, the so-called soft output of the classifier $\nu: \mathbb{X}\mapsto \left [0,1\right ]^{2}$ is also defined. The soft output vector $\nu$ contains values   proportional to the conditional probabilities. Consequently, the following conditions need to be satisfied:
\begin{align}
&\nu_{i} \approx P(\mathbb{M}=i|\mathbb{X}=x),\\
&\nu_{i}(\vec{x}) \in \left [ 0 , 1 \right ], \\
&\sum_{i=1}^{M}\nu_{i}(\vec{x}) = 1.
\end{align}

\subsection{Soft Confusion Matrix}\label{sec:PropMethod:SCM}
The SCM approach is based on an assessment of the probability of classifying an object $\vec{x}$ into the class $s \in \mathbb{M}$  using the classifier $\psi$. It also provides an extension of the Bayesian model in which the object’s description $\vec{x}$ and its true label $m \in \mathbb{M}$ are realizations of random variables  $\vec{\textbf{X}}$ and $\textbf{M}$, respectively.  In the SCM approach, classifier $\psi$ predicts randomly based on the probabilities $P(\mathbf{\Psi}(\vec{x})=s)=P(s|\vec{x})$~\cite{Berger1985}. Hence, the outcome of the classification $s$ is a~realization of the random variable $\mathbf{\Psi}(\vec{x})$. Unfortunately, for deterministic classifiers, these probabilities would be zero or one. The problem may be dealt with using a randomized classifier equivalent to the given one (RRC). 

According to the extended Bayesian model, the posterior probability $P(m|\vec{x})$ of label $m$ can be defined as:
\begin{align}\label{eq:postProb1}
 P(m|\vec{x}) &= \sum_{s \in \mathcal{M}} P(s|\vec{x}) P(m|s,\vec{x}). 
\end{align}
where $P(m|s,\vec{x})$ denotes the probability that an object $\vec{x}$ belongs to the class $m$ given that $\mathbf{\Psi}(\vec{x})=s$. This probability is estimated using local soft confusion matrix. The locality of the matrix is defined using Gaussian potential function with $\beta$ parameter. The detailed procedure of obtainging this matrix is given in~\cite{Trajdos2016}.

Unfortunately, the assumption that base classifier assigns labels in a~stochastic way is rather impractical, since most real-life classifiers are deterministic. This issue was addressed by implementation of deterministic binary classifiers in which their statistical properties were modelled using the RRC procedure, as described in section~\ref{sec:PropMethod:RRC}. 

\subsection{Randomized Reference Classifier}\label{sec:PropMethod:RRC}
In above-mentioed approach, the behaviour of a~base classifier $\psi$ was modeled using a~stochastic classifier defined by a~probability distribution over the set of labels $\mathbb{M}$. In this study, the randomized reference classifier (RRC) proposed by Woloszynski and Kurzynski~\cite{Woloszynski2011} was used. The RRC is a~hypothetical classifier that allows a~randomised model of a~given deterministic classifier to be built.

We assumed that for a~given instance $\vec{x}$, the randomised classifier $\psi^{(R)} $ generates a~vector of class supports  $\nu$ being observed values of random variables $\Delta_{i}(\vec{x})$. The chosen probability distribution of random variables needs to satisfy the following conditions:
\begin{align}
\label{MK_PT:delta_c1}
\Delta_{i}(\vec{x}) &\in [0,1], \\
\label{MK_PT:delta_c2}
\sum_{i=1}^{M}\Delta_{i}(\vec{x})&=1, \\
\label{MK_PT:delta_c3}
\mathbf{E}\left[\Delta_{i}(\vec{x}) \right] &= \nu_{i}(\vec{x}),\ i \in \{0,1 \},
\end{align}
where $\mathbf{E}$ is the expected value operator. Conditions (\ref{MK_PT:delta_c1})  and (\ref{MK_PT:delta_c2})  follow from the normalisation properties of class supports, whereas condition (\ref{MK_PT:delta_c3}) provides the equivalence of the randomized model $\psi^{(R)}$ and base classifier $\psi$. Based on the latter condition, the RRC can be used to provide a~randomised model of any classifier that returns a~vector of class-specific supports $\nu(\vec{x})$. 

The probability of classifying an object $\vec{x}$ into the class $i$ using the RRC can be calculated from the following formula:
\begin{equation}   \label{MK_PT:wzor5}
P(\mathbf{\Psi}=m|\vec{\textbf{X}}=\vec{x})=Pr\left[\Delta_{m}(\vec{x})> \Delta_{\mathbb{M} \setminus m}(\vec{x})\right],
\end{equation}
where $Pr\left[\Delta_{m}(\vec{x})> \Delta_{\mathbb{M} \setminus m}(\vec{x})\right]$ is the probability that the value obtained by the realisation of random variable $\Delta_{m}$ is greater than the realisation of the remaining random variables.

In this paper, we propose using the normal distribution truncated to the inverval $[0,1]$~\cite{johnson1994continuous} instead of the beta distribution suggested by Woloszynski et al.~\cite{Woloszynski2011}. The expected value for each random variable is simply determined using formula~\eqref{MK_PT:delta_c3}. The standard deviation is determined using a rescaled variance of the beta distribution:
\begin{align}\label{eq:stdDev}
 sd_{i} &= \left(\frac{\nu_{i}(1-\nu_{i})}{M+1}\right)^{\normSDParam},
\end{align}
where $\normSDParam$ is a parameter that should be tuned in order to achieve the best classification quality of the SCM method. 

\section{Experimental Setup}\label{sec:ExSetup}

The goal of this paper is to determine how changing the underlying distribution of RRC classifier affect the classification quality of algorithms built using RRC model. To do so the experimental evaluation, which setup is described below, is performed. 

The following base classifiers were employed:
\begin{itemize}
 \item $\psi_{\mathrm{NB}}$ -- Naive Bayes classifier with kernel density estimation~\cite{Hand2001}.
 \item $\psi_{\mathrm{KNN}}$ --  nearest neighbours classifier~\cite{Cover1967}.
 \item $\psi_{\mathrm{J48}}$ -- Weka implementation of the C4.5 algorithm~\cite{Quinlan1993}. Laplace smoothing is used to produce estimation of conditional probability~\cite{Provost2003}
 \item $\psi_{\mathrm{NC}}$ -- nearest centroid (Nearest Prototype)~\cite{Kuncheva1998}
\end{itemize}
The classifiers implemented in WEKA framework~\cite{Hall2009} were used. If not stated otherwise, the classifier parameters were set to their defaults. For the KNN classifier, the number of neighbours was selected from the following values $K \in \{1,3,5,\ldots, 11 \}$.

During the experimental evaluation the following classifiers  were compared: 
\begin{enumerate}
 \item $\psi_{\mathrm{R}}$ -- unmodified base classifier,
 \item $\psi_{\mathrm{B}}$ -- SCM classifier with beta distribution,
 \item $\psi_{\mathrm{N}}$ -- SCM classifier with truncated normal distribution.
\end{enumerate}

The size of the neighborhood, expressed as $\beta$ coefficient and the variance-related coefficient $\normSDParam$, were chosen using a~fiveefold cross-validation procedure and the grid search technique. The following values of $\beta$ and $\normSDParam$ were considered: $ \beta \in \left\{1,2,3, \cdots, 21 \right\}$, $\normSDParam \in \left\{0.1,0.2,0.3,\cdots,1.0\right\}$. The values were chosen in such a way that minimizes macro-averaged $F_1$ loss function.

The experimental code was implemented using WEKA framework~\cite{Hall2009}.The source code of the algorithms is available online~\footnote{\url{https://github.com/ptrajdos/rrcBasedClassifiers/tree/develop}}. 

To evaluate the proposed methods the following classification-loss criteria are used~\cite{Sokolova2009}: Zero-one loss (1-Accuracy); Macro-averaged $\fdr$ (1- precision), $\fnr$ (1-recall), $\fOne$;Micro-averaged $\fdr$, $\fnr$, $\fOne$. 

Following the recommendations of~\cite{demsar2006} and~\cite{garcia2008extension}, the statistical significance of the obtained results was assessed using the two-step procedure. The first step is to perform the Friedman test~\cite{Friedman1940} for each quality criterion separately. Since the multiple criteria were employed, the familywise errors (\FWER{}) should be controlled~\cite{benjamini2001control}. To do so, the Bergman-Hommel~\cite{Bergmann1988} procedure of controlling \FWER{} of the conducted Friedman tests was employed. When the Friedman test shows that there is a significant difference within the group of classifiers, the pairwise tests using the Wilcoxon signed-rank test~\cite{wilcoxon1945,demsar2006} were employed. To control \FWER{} of the Wilcoxon-testing procedure, the Bergman-Hommel approach was employed~\cite{Bergmann1988}. For all tests the significance level was set to $\alpha=0.05$.

Table~\ref{tab:BenchmarkSetsCharacteristics} displays the collection of the $64$ benchmark sets that were used during the experimental evaluation of the proposed algorithms. The table is divided into three columns.  Each column is organized as follows. The first column contains the names of the datasets. The remaining ones contain the set-specific characteristics of the benchmark sets: The number of instances in the dataset ($|S|$); dimensionality of the input space ($d$); the number of classes ($C$);average imbalance ratio ($\mathrm{IR}$). Benchmark datasets are available online~\footnote{\url{https://github.com/ptrajdos/MLResults/blob/master/data/slDataFull.zip}}.

To reduce the computational burden and remove irrelevant features, the correlation-based feature selection described in~\cite{Hall1999} was applied. 

{
\setlength\tabcolsep{2.0pt}%
\def\arraystretch{0.9}%
\begin{table}
 \centering\tiny
 \caption{The characteristics of the benchmark sets}\label{tab:BenchmarkSetsCharacteristics}
 \begin{tabular}{lcccc|lcccc|lcccc}
Name&$|S|$&$d$&$C$&$\mathrm{IR}$&Name&$|S|$&$d$&$C$&$\mathrm{IR}$&Name&$|S|$&$d$&$C$&$\mathrm{IR}$\\
\hline
appendicitis&106&7&2&2.52&	housevotes&435&16&2&1.29&	shuttle&57999&9&7&1326.03\\
australian&690&14&2&1.12&	ionosphere&351&34&2&1.39&	sonar&208&60&2&1.07\\
balance&625&4&3&2.63&	iris&150&4&3&1.00&	spambase&4597&57&2&1.27\\
banana2D&2000&2&2&1.00&	led7digit&500&7&10&1.16&	spectfheart&267&44&2&2.43\\
bands&539&19&2&1.19&	lin1&1000&2&2&1.01&	spirals1&2000&2&2&1.00\\
Breast Tissue&105&9&6&1.29&	lin2&1000&2&2&1.83&	spirals2&2000&2&2&1.00\\
check2D&800&2&2&1.00&	lin3&1000&2&2&2.26&	spirals3&2000&2&2&1.00\\
cleveland&303&13&5&5.17&	magic&19020&10&2&1.42&	texture&5500&40&11&1.00\\
coil2000&9822&85&2&8.38&	mfdig fac&2000&216&10&1.00&	thyroid&7200&21&3&19.76\\
dermatology&366&34&6&2.41&	movement libras&360&90&15&1.00&	titanic&2201&3&2&1.55\\
diabetes&768&8&2&1.43&	newthyroid&215&5&3&3.43&	twonorm&7400&20&2&1.00\\
Faults&1940&27&7&4.83&	optdigits&5620&62&10&1.02&	ULC&675&146&9&2.17\\
gauss2DV&800&2&2&1.00&	page-blocks&5472&10&5&58.12&	vehicle&846&18&4&1.03\\
gauss2D&4000&2&2&1.00&	penbased&10992&16&10&1.04&	Vertebral Column&310&6&3&1.67\\
gaussSand2&600&2&2&1.50&	phoneme&5404&5&2&1.70&	wdbc&569&30&2&1.34\\
gaussSand&600&2&2&1.50&	pima&767&8&2&1.44&	wine&178&13&3&1.23\\
glass&214&9&6&3.91&	ring2D&4000&2&2&1.00&	winequality-red&1599&11&6&20.71\\
haberman&306&3&2&1.89&	ring&7400&20&2&1.01&	winequality-white&4898&11&7&82.94\\
halfRings1&400&2&2&1.00&	saheart&462&9&2&1.44&	wisconsin&699&9&2&1.45\\
halfRings2&600&2&2&1.50&	satimage&6435&36&6&1.66&	yeast&1484&8&10&17.08\\
hepatitis&155&19&2&2.42&	Seeds&210&7&3&1.00&&&&&\\
HillVall&1212&100&2&1.01&	segment&2310&19&7&1.00&&&&&\\
\end{tabular}
\end{table}
}

\section{Results and Discussion}\label{sec:ResAndDisc}

\begin{figure}[tb]
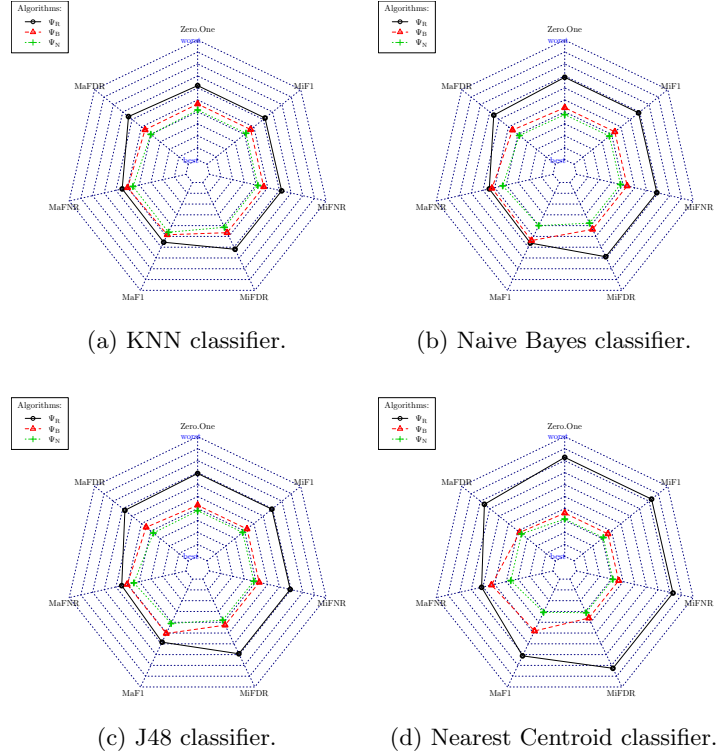

\begin{center}

\subfloat[KNN classifier.]{\label{fig:radarKNN}%
  \includegraphics[width=0.4\textwidth]{\ptFiguresDirectory{radarKNN}}%
}%
\subfloat[Naive Bayes classifier.]{\label{fig:radarNB}%
  \includegraphics[width=0.4\textwidth]{\ptFiguresDirectory{radarNB}}%
}

\subfloat[J48 classifier.]{\label{fig:radarJ48}%
  \includegraphics[width=0.4\textwidth]{\ptFiguresDirectory{radarJ48}}%
}%
\subfloat[Nearest Centroid classifier.]{\label{fig:radarNC}%
  \includegraphics[width=0.4\textwidth]{\ptFiguresDirectory{radarNC}}%
}
\caption{Radar plots for the investigated classifiers.} 
\end{center}
\end{figure}

To compare multiple algorithms on multiple benchmark sets the average ranks approach~\cite{demsar2006} is used. In the approach, the winning algorithm achieves rank equal '1', the second achieves rank equal '2', and so on. In the case of ties, the ranks of algorithms that achieve the same results, are averaged. To provide a visualisation of the average ranks, the radar plots are employed. In the plots, the data is visualised in such way that the lowest ranks are closer to the centre of the graph. The radar plots related to the experimental results are shown in figures~\ref{fig:radarKNN} -- \ref{fig:radarNC}.

Due to the page limit, the full results are published online~\footnote{\url{https://github.com/ptrajdos/MLResults/blob/master/Boundaries/bounds_hetero_15.01.2019E4_m_R.zip}} 

The numerical results are given in Table~\ref{table:KNNStat}.The table is structured as follows. The table is divided into-base-classifier-specific sections and each section has it's own header containing base classifier name. The first row of each section contains names of the investigated algorithms. Then the table is divided into seven sections -- one section is related to a single evaluation criterion. The first row of each section is the name of the quality criterion investigated in the section. The second row shows the p-value of the Friedman test. The third one shows the average ranks achieved by algorithms. The following rows show p-values resulting from pairwise Wilcoxon test. The p-value equal to $0.000$ informs that the p-values are lower than $10^{-3}$ and p-value equal to $1.000$ informs that the value is higher than $0.999$.

Let us begin with the analysis of results for zero-one loss and micro-averaged criteria which are known to be biased towards majority classes~\cite{Sokolova2009}. For all investigated base classifiers the results are pretty consistent. That is, SCM-based classifiers are significantly better than the unmodified classifier. And there are no significant differences between $\psi_{\mathrm{B}}$ and $\psi_{\mathrm{N}}$ classifiers. 

For macro-averaged criteria, on the other hand, the results are a bit different. Generally, the average ranks suggest that truncated-normal-based SCM classifiers may be a bit better than the beta-based SCM. However, not all differences are significant. For $\psi_{\mathrm{KNN}}$ classifier there are no significant differences between the investigated methods. For $\psi_{\mathrm{NB}}$ and $\psi_{\mathrm{J48}}$ SCM-based classifiers are significantly better according to $\fdr$ criterion.  For $\psi_{\mathrm{NC}}$ base classifier (the weakest one) the $\psi_{\mathrm{N}}$ classifier outperforms the remaining classifiers. These results suggest that truncated-normal-based SCM classifier is a bit better than beta-based SCM method in discovering objects coming from minority classes. 

{
\setlength\tabcolsep{1.0pt}%
\begin{table}[htb]
\centering\tiny
\caption{Statistical evaluation. Wilcoxon test results.\label{table:KNNStat}}
\begin{tabular}{c|ccc|ccc|ccc|ccc|ccc|ccc|ccc}
\multicolumn{22}{c}{$\psi_{\mathrm{KNN}}$}\\
  & $\psi_{\mathrm{R}}$ & $\psi_{\mathrm{B}}$ & $\psi_{\mathrm{N}}$ & $\psi_{\mathrm{R}}$ & $\psi_{\mathrm{B}}$ & $\psi_{\mathrm{N}}$ & $\psi_{\mathrm{R}}$ & $\psi_{\mathrm{B}}$ & $\psi_{\mathrm{N}}$ & $\psi_{\mathrm{R}}$ & $\psi_{\mathrm{B}}$ & $\psi_{\mathrm{N}}$ & $\psi_{\mathrm{R}}$ & $\psi_{\mathrm{B}}$ & $\psi_{\mathrm{N}}$ & $\psi_{\mathrm{R}}$ & $\psi_{\mathrm{B}}$ & $\psi_{\mathrm{N}}$ & $\psi_{\mathrm{R}}$ & $\psi_{\mathrm{B}}$ & $\psi_{\mathrm{N}}$ \\ 
  \hline
Nam&\multicolumn{3}{c|}{Zero-One}&\multicolumn{3}{c|}{MaFDR}&\multicolumn{3}{c|}{MaFNR}&\multicolumn{3}{c|}{MaF1}&\multicolumn{3}{c|}{MiFDR}&\multicolumn{3}{c|}{MiFNR}&\multicolumn{3}{c}{MiF1}\\
Frd&\multicolumn{3}{c|}{7.667e-02}&\multicolumn{3}{c|}{2.360e-02}&\multicolumn{3}{c|}{8.487e-01}&\multicolumn{3}{c|}{8.487e-01}&\multicolumn{3}{c|}{7.667e-02}&\multicolumn{3}{c|}{7.667e-02}&\multicolumn{3}{c}{7.667e-02}\\
Rnk & 2.24 & 1.93 & 1.83 & 2.28 & 1.92 & 1.80 & 2.09 & 2.00 & 1.91 & 2.11 & 1.97 & 1.92 & 2.24 & 1.93 & 1.83 & 2.24 & 1.93 & 1.83 & 2.24 & 1.93 & 1.83 \\ 
   \cmidrule(lr){2-4}\cmidrule(lr){5-7}\cmidrule(lr){8-10}\cmidrule(lr){11-13}\cmidrule(lr){14-16}\cmidrule(lr){17-19}\cmidrule(lr){20-22}
$\psi_{\mathrm{R}}$ &  & .043 & .043 &  & .236 & .236 &  & 1.000 & 1.000 &  & 1.000 & 1.000 &  & .043 & .043 &  & .043 & .043 &  & .043 & .043 \\ 
  $\psi_{\mathrm{B}}$ &  &  & .432 &  &  & .270 &  &  & 1.000 &  &  & 1.000 &  &  & .432 &  &  & .432 &  &  & .432 \\ 
  
  \multicolumn{22}{c}{$\psi_{\mathrm{NB}}$}\\
  & $\psi_{\mathrm{R}}$ & $\psi_{\mathrm{B}}$ & $\psi_{\mathrm{N}}$ & $\psi_{\mathrm{R}}$ & $\psi_{\mathrm{B}}$ & $\psi_{\mathrm{N}}$ & $\psi_{\mathrm{R}}$ & $\psi_{\mathrm{B}}$ & $\psi_{\mathrm{N}}$ & $\psi_{\mathrm{R}}$ & $\psi_{\mathrm{B}}$ & $\psi_{\mathrm{N}}$ & $\psi_{\mathrm{R}}$ & $\psi_{\mathrm{B}}$ & $\psi_{\mathrm{N}}$ & $\psi_{\mathrm{R}}$ & $\psi_{\mathrm{B}}$ & $\psi_{\mathrm{N}}$ & $\psi_{\mathrm{R}}$ & $\psi_{\mathrm{B}}$ & $\psi_{\mathrm{N}}$ \\ 
  \hline
Nam&\multicolumn{3}{c|}{Zero-One}&\multicolumn{3}{c|}{MaFDR}&\multicolumn{3}{c|}{MaFNR}&\multicolumn{3}{c|}{MaF1}&\multicolumn{3}{c|}{MiFDR}&\multicolumn{3}{c|}{MiFNR}&\multicolumn{3}{c}{MiF1}\\
Frd&\multicolumn{3}{c|}{1.960e-03}&\multicolumn{3}{c|}{9.098e-03}&\multicolumn{3}{c|}{3.327e-01}&\multicolumn{3}{c|}{2.205e-01}&\multicolumn{3}{c|}{1.960e-03}&\multicolumn{3}{c|}{1.960e-03}&\multicolumn{3}{c}{1.960e-03}\\
Rnk & 2.38 & 1.87 & 1.75 & 2.31 & 1.92 & 1.77 & 2.09 & 2.05 & 1.86 & 2.12 & 2.08 & 1.80 & 2.38 & 1.87 & 1.75 & 2.38 & 1.87 & 1.75 & 2.38 & 1.87 & 1.75 \\ 
   \cmidrule(lr){2-4}\cmidrule(lr){5-7}\cmidrule(lr){8-10}\cmidrule(lr){11-13}\cmidrule(lr){14-16}\cmidrule(lr){17-19}\cmidrule(lr){20-22}
$\psi_{\mathrm{R}}$ &  & .000 & .000 &  & .003 & .003 &  & .183 & .069 &  & .089 & .083 &  & .000 & .000 &  & .000 & .000 &  & .000 & .000 \\ 
  $\psi_{\mathrm{B}}$ &  &  & .151 &  &  & .131 &  &  & .069 &  &  & .083 &  &  & .152 &  &  & .152 &  &  & .152 \\ 
  
  \multicolumn{22}{c}{$\psi_{\mathrm{J48}}$}\\
  & $\psi_{\mathrm{R}}$ & $\psi_{\mathrm{B}}$ & $\psi_{\mathrm{N}}$ & $\psi_{\mathrm{R}}$ & $\psi_{\mathrm{B}}$ & $\psi_{\mathrm{N}}$ & $\psi_{\mathrm{R}}$ & $\psi_{\mathrm{B}}$ & $\psi_{\mathrm{N}}$ & $\psi_{\mathrm{R}}$ & $\psi_{\mathrm{B}}$ & $\psi_{\mathrm{N}}$ & $\psi_{\mathrm{R}}$ & $\psi_{\mathrm{B}}$ & $\psi_{\mathrm{N}}$ & $\psi_{\mathrm{R}}$ & $\psi_{\mathrm{B}}$ & $\psi_{\mathrm{N}}$ & $\psi_{\mathrm{R}}$ & $\psi_{\mathrm{B}}$ & $\psi_{\mathrm{N}}$ \\ 
  \hline
Nam&\multicolumn{3}{c|}{Zero-One}&\multicolumn{3}{c|}{MaFDR}&\multicolumn{3}{c|}{MaFNR}&\multicolumn{3}{c|}{MaF1}&\multicolumn{3}{c|}{MiFDR}&\multicolumn{3}{c|}{MiFNR}&\multicolumn{3}{c}{MiF1}\\
Frd&\multicolumn{3}{c|}{4.727e-04}&\multicolumn{3}{c|}{7.379e-04}&\multicolumn{3}{c|}{3.987e-01}&\multicolumn{3}{c|}{1.646e-01}&\multicolumn{3}{c|}{4.727e-04}&\multicolumn{3}{c|}{4.727e-04}&\multicolumn{3}{c}{4.727e-04}\\
Rnk & 2.38 & 1.85 & 1.76 & 2.35 & 1.90 & 1.75 & 2.10 & 2.01 & 1.89 & 2.17 & 2.01 & 1.82 & 2.38 & 1.85 & 1.76 & 2.38 & 1.85 & 1.76 & 2.38 & 1.85 & 1.76 \\ 
   \cmidrule(lr){2-4}\cmidrule(lr){5-7}\cmidrule(lr){8-10}\cmidrule(lr){11-13}\cmidrule(lr){14-16}\cmidrule(lr){17-19}\cmidrule(lr){20-22}
$\psi_{\mathrm{R}}$ &  & .000 & .000 &  & .010 & .003 &  & .220 & .220 &  & .118 & .118 &  & .000 & .000 &  & .000 & .000 &  & .000 & .000 \\ 
  $\psi_{\mathrm{B}}$ &  &  & .078 &  &  & .206 &  &  & .220 &  &  & .118 &  &  & .078 &  &  & .078 &  &  & .078 \\ 
  
  \multicolumn{22}{c}{$\psi_{\mathrm{NC}}$}\\
   & $\psi_{\mathrm{R}}$ & $\psi_{\mathrm{B}}$ & $\psi_{\mathrm{N}}$ & $\psi_{\mathrm{R}}$ & $\psi_{\mathrm{B}}$ & $\psi_{\mathrm{N}}$ & $\psi_{\mathrm{R}}$ & $\psi_{\mathrm{B}}$ & $\psi_{\mathrm{N}}$ & $\psi_{\mathrm{R}}$ & $\psi_{\mathrm{B}}$ & $\psi_{\mathrm{N}}$ & $\psi_{\mathrm{R}}$ & $\psi_{\mathrm{B}}$ & $\psi_{\mathrm{N}}$ & $\psi_{\mathrm{R}}$ & $\psi_{\mathrm{B}}$ & $\psi_{\mathrm{N}}$ & $\psi_{\mathrm{R}}$ & $\psi_{\mathrm{B}}$ & $\psi_{\mathrm{N}}$ \\ 
  \hline
Nam&\multicolumn{3}{c|}{Zero-One}&\multicolumn{3}{c|}{MaFDR}&\multicolumn{3}{c|}{MaFNR}&\multicolumn{3}{c|}{MaF1}&\multicolumn{3}{c|}{MiFDR}&\multicolumn{3}{c|}{MiFNR}&\multicolumn{3}{c}{MiF1}\\
Frd&\multicolumn{3}{c|}{1.750e-09}&\multicolumn{3}{c|}{4.146e-06}&\multicolumn{3}{c|}{1.263e-02}&\multicolumn{3}{c|}{3.022e-05}&\multicolumn{3}{c|}{1.750e-09}&\multicolumn{3}{c|}{1.750e-09}&\multicolumn{3}{c}{1.750e-09}\\
Rnk & 2.65 & 1.72 & 1.62 & 2.52 & 1.76 & 1.72 & 2.22 & 2.05 & 1.72 & 2.42 & 1.96 & 1.61 & 2.65 & 1.72 & 1.62 & 2.65 & 1.72 & 1.62 & 2.65 & 1.72 & 1.62 \\ 
   \cmidrule(lr){2-4}\cmidrule(lr){5-7}\cmidrule(lr){8-10}\cmidrule(lr){11-13}\cmidrule(lr){14-16}\cmidrule(lr){17-19}\cmidrule(lr){20-22}
$\psi_{\mathrm{R}}$ &  & .000 & .000 &  & .000 & .000 &  & .440 & .055 &  & .019 & .000 &  & .000 & .000 &  & .000 & .000 &  & .000 & .000 \\ 
  $\psi_{\mathrm{B}}$ &  &  & .102 &  &  & .453 &  &  & .003 &  &  & .002 &  &  & .102 &  &  & .102 &  &  & .102 \\ 
  \end{tabular}
\end{table}
}

\section{Conclusions}\label{sec:Conc}

In this paper, the issue of building the RRC model using truncated-normal distribution has been investigated. During the experimental evaluation, promising results have been obtained. Despite a naive-heuristic method has been applied to estimate the variance of the underlying normal distribution, the proposed method is comparable to the original beta-distribution-based approach. What is more, for some classifiers the results show that the proposed approach is better at discovering objects coming from minority classes. We believe that applying a better method of variance estimation will improve the results. Consequently, our further research will explore this issue. 

\subsubsection*{Acknowledgments.} This work was supported by the statutory funds of the Department of Systems and Computer Networks, Wroclaw University of Science and Technology.

\bibliography{bibliography}

\end{document}